\documentclass{article}

\usepackage{arxiv}

\usepackage[utf8]{inputenc} 
\usepackage[T1]{fontenc}    
\usepackage{hyperref}       
\usepackage{url}            
\usepackage{booktabs}       
\usepackage{amsfonts}       
\usepackage{nicefrac}       
\usepackage{microtype}      
\usepackage{lipsum}		
\usepackage{graphicx}
\usepackage{subcaption}
\usepackage[square,sort,comma,numbers]{natbib}
\usepackage{amssymb}
\usepackage{pifont}

\newcommand{\cmark}{\ding{51}}%
\newcommand{\xmark}{\ding{55}}%

\title{Working with scale: 2nd place solution to Product Detection in Densely Packed Scenes
}

\author{\hspace{1pt}{\hspace{1pt}Artem Kozlov}\\
	\texttt{imtyommy@gmail.com} \\
}

\date{} 					

\renewcommand{\headeright}{Technical Report}
\renewcommand{\undertitle}{Technical Report}

\hypersetup{
pdftitle={Working with scale: 2nd place solution to Product Detection in Densely Packed Scenes},
pdfsubject={},
pdfauthor={Artem Kozlov},
pdfkeywords={},
}

\begin{document}
\maketitle

\begin{abstract}
	This report describes a 2nd place solution of the \emph{detection challenge} which is held within CVPR 2020 Retail-Vision workshop. Instead of going further considering previous results this work mainly aims to verify previously observed takeaways by re-experimenting. The reliability and reproducibility of the results are reached by incorporating a popular object detection toolbox - \emph{MMDetection}. In this report, I  firstly represent the results received for Faster-RCNN and RetinaNet models, which were taken for comparison in the original work. Then I describe the experiment results with more advanced models. The final section reviews two simple tricks for Faster-RCNN model that were used for my final submission: changing default anchor scale parameter and train-time image tiling. The source code is available at \url{https://github.com/tyomj/product_detection}.
\end{abstract}


\section{Introduction}
\label{sec:intro}

The majority of works in object detection mainly describe results on two datasets: Pascal VOC \cite{everingham2011pascal} and MS-COCO \cite{lin2014microsoft}. While this information might be helpful in forming the relative intuition of a model behavior for common scenes, it is still unobvious when it comes to a dataset whose distribution of objects deviates. Goldman et al. \cite{goldman2019precise} proposed a dataset with the following key features: 1) a high number of objects on an image, 2) objects are densely packed.

Although in the comparison table the authors declared the dataset to have 110,712 classes, it is important to note that in this context they used the term \emph{class} to describe the appearance variations among the objects. De facto the dataset represents a single-class object detection task. Considering this information, we might reference an example of a similar task from a different domain \cite{shao2018crowdhuman} where both RetinaNet \cite{lin2017focal} and Faster-RCNN \cite{ren2015faster} were taking as baseline solutions and showed good results. Hence, the first step of my plan was to revisit some of the baseline architectures.

A rapidly growing number of papers with different new methods in object detection makes it difficult to maintain the reliable and up-to-date benchmarking table especially when a custom implementation of the new method also comes with different parts of the training pipeline such as data pre-processing or results post-processing. In order to address this issue, many attempts have been made for the last several years to create a highly reusable codebase that covers most of the popular models in the field. For now there are plenty of such toolboxes based on different deep learning frameworks: Tensorflow Object Detection API \cite{huang2016speedaccuracy}, Detectron \cite{Detectron2018} and Detectron2 \cite{wu2019detectron2}, MMDetection \cite{chen2019mmdetection}, and so on. Each of these toolboxes has reusable parts of code, which makes it easier to compare the performance of models in vitro. For my experiments I chose MMDetection v1.

The rest of the report is structured as follows: in Section \ref{sec:dataset} I briefly review the parameters of the dataset and additionally highlight the features which I found important. Section  \ref{sec:exps} 
contains all the experiment results where I consequentially inspect the results starting from baseline models \ref{sec:basleine} then testing a few approaches that are different from widely-used anchoring scheme \ref{sec:adv}. In the final subsection \ref{sec:scale} I describe two simple tricks that helped to achieve considerably good performance for Faster-RCNN, testing some commonly-used tricks along the way.

\section{Dataset description}
\label{sec:dataset}

The dataset for the challenge is almost the same version that was originally described in \cite{goldman2019precise} with minor differences: the train set contains 8,219 instead of 8,233 images and the test set contains 2,936 out of 2,941. The validation part contains exactly 588 images. During training a few more examples of broken images were found, so those might be easily excluded. 
\begin{figure}[h!]
  \centering
  \begin{subfigure}[b]{0.4\linewidth}
    \includegraphics[width=\linewidth]{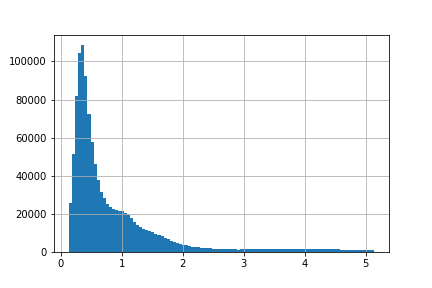}
    \caption{Aspect ratio}
  \end{subfigure}
  \begin{subfigure}[b]{0.4\linewidth}
    \includegraphics[width=\linewidth]{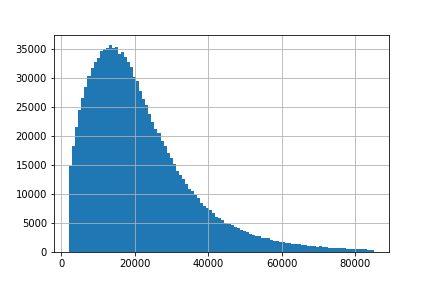}
    \caption{Area}
  \end{subfigure}
  \begin{subfigure}[b]{0.4\linewidth}
    \includegraphics[width=\linewidth]{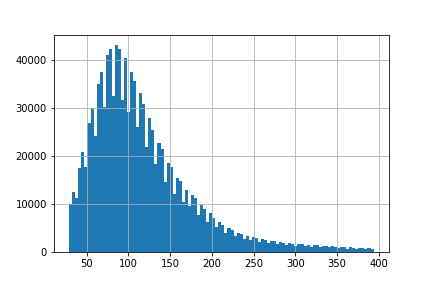}
    \caption{Width}
  \end{subfigure}
  \begin{subfigure}[b]{0.4\linewidth}
    \includegraphics[width=\linewidth]{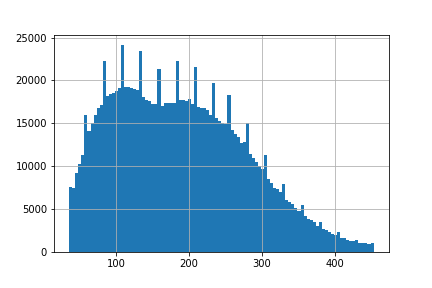}
    \caption{Height}
  \end{subfigure}
  \caption{Statistics of object geometry in histograms. Outliers are clipped by 0.01 and 0.99 quantiles.}
  \label{fig:fig1}
\end{figure}

Inspecting the train set statistics the most popular image sizes were found. It turned out that the majority of scenes were taken in portrait mode in the following resolutions: 2448x3264 - 3463 images, 1920x2560
- 1710 images, 2336x4160 and 2340x4160 might be combined to have 1133 samples in total. The distribution of image sizes definitely differs from Pascal VOC and MS COCO that is a premise to experimenting with related parameters. Secondly, object sizes and aspect ratios were reviewed (see Fig. \ref{fig:fig1}). While aspect ratio distribution was not found interesting, object size distribution seemed a more rewarding feature to work with. Although object sizes were distributed as 0.12\%/18.53\%/81.35\% among "small", "medium" and "large" according to MS COCO the object size classification, the picture is changing in the scenario of image resizing, which is a necessary trick for training with reasonable batch size. Decreasing the image size to 816x1088 \footnote{This size is used as a default parameter to "Resize" transformation for the rest of the report until otherwise is specified. Unlike default 1333x800 the proposed size takes into account the most popular image size within SKU110k.} which is 9 times smaller, we will see the following shift - 18.64\%/80.27\%/1.09\%. In this scale the number of "large" examples is substantially small. These findings drew much attention to the ideas described in \ref{sec:scale}.

\section{Experiments}
\label{sec:exps}

\subsection{Previously compared models}
\label{sec:basleine}
For the purpose of setting up a fair baseline two main models were chosen: Faster-RCNN and RetinaNet. Both models are used with default configurations utilizing ResNet-50 \cite{he2015deep} backbone and FPN \cite{lin2016feature}. In the initial experiments images are resized to a maximum scale of 1333x800, without changing the aspect ratio. The training schedule is “1x”, which means 12 epochs. "RandomFlip" image augmentation is used with probability 0.5 during training time. The model is trained on the training set, and the test set is used for evaluation. Evaluation is performed using official "pycocotools" package which is an implementation of detection evaluation metrics used by COCO with "maxDets" parameter changed to "[1 10 300]". \footnote{The results obtained by testing with pycocotools were slightly different from the results obtained with the evaluation code proposed by the authors. I did not set a goal of revealing the differences in code, so just be aware while comparing with any results taken from other technical reports.}

Given the results described in Table \ref{tab:table1} I decided to further experiment with stronger models.

\begin{table}
	\caption{Revisiting default models}
	\centering
	\begin{tabular}{llllll}
		\toprule
		Model               &  mAP  & AP@$^{0.5}$ & AP@$^{0.75}$ & AR$^{300}$   \\
		\midrule
		RetinaNet-r50-fpn    & 0.463 & 0.751  & 0.532 & 0.512 \\
		Faster-RCNN-r50-fpn  & 0.523 & 0.850  & 0.592 & 0.582 \\
		\bottomrule
	\end{tabular}
	\label{tab:table1}
\end{table}

\subsection{Different anchoring approaches}
\label{sec:adv}

In order to test anchoring strategies different from the dense scheme used for both previous models \emph{Guided Anchoring} \cite{wang2019region} was tried in a combination with RetinaNet on top of two different backbones. Surprisingly, GA-RetinaNet-r50-fpn showed results comparable with Faster-RCNN model. Another approach \emph{RepPoints} \cite{yang2019reppoints}, which is an example of anchor free detectors, did not show any significant improvement with default parameters.

\begin{table}[!htp]
	\caption{Non-dense anchoring}
	\centering
	\begin{tabular}{llllll}
		\toprule
		Model                        &  mAP  & AP@$^{0.5}$ & AP@$^{0.75}$ & AR$^{300}$   \\
		\midrule
		GA-RetinaNet-r50-fpn         & 0.523 & 0.870 & 0.579 & 0.583 \\
		GA-RetinaNet-x101-32x4d-fpn  & 0.537 & 0.882 & 0.602 & 0.598 \\
		RepPoints-moment-r50-fpn     & 0.505 & 0.815 & 0.578 & 0.562 \\
		\bottomrule
	\end{tabular}
	\label{tab:table2}
\end{table}

\subsection{Scale-related tricks}
\label{sec:scale}

\subsubsection{Anchor scale}
\label{sec:scale1}
MMDetection Faster-RCNN default implementation uses the following anchor settings: `anchor\_strides`=[4, 8, 16, 32, 64] that specifies the stride for an anchor grid and `anchor\_scale` = [8] that is used as a multiplication factor for strides to obtain 5 anchor scales with areas [32$^2$, 64$^2$, 128$^2$, 256$^2$, 512$^2$] respectively. In Section \ref{sec:dataset} I highlighted lack of "large" objects if resizing preprocessing is applied. Hence, in order to generate more dense proposals, it looked like a good idea to decrease `anchor\_scale` to [4] to form the following areas [16$^2$, 32$^2$, 64$^2$, 128$^2$, 256$^2$]. 
Table \ref{tab:table3} represent relative importance of this parameter.

\begin{table}[!htp]
	\caption{Comparison of different anchor scales}
	\centering
	\begin{tabular}{llllll}
		\toprule
		Model                &  Anchor scale &  mAP  & AP@$^{0.5}$ & AP@$^{0.75}$ & AR$^{300}$   \\
		\midrule
		Faster-RCNN-r50-fpn  & [8] & 0.522	&0.850	&0.591	&0.577 \\
		Faster-RCNN-r50-fpn  & [4] & 0.551	&0.912	&0.614	&0.613 \\
		\bottomrule
	\end{tabular}
	\label{tab:table3}
\end{table}

To pay respect to computer vision challenges widely-used tricks were tested on top of that such as: simple TTA with `RandomFlip`, additional data augmentation using albumentations \cite{info11020125} config\footnote{ \url{https://github.com/open-mmlab/mmdetection/blob/v1.0.0/configs/albu_example/mask_rcnn_r50_fpn_1x.py}}, multi-scale training. None of them gave any significant boost. See Table \ref{tab:table4}.

\begin{table}[!htp]
	\caption{Some of the common tricks}
	\centering
	\begin{tabular}{llllllllll}
		\toprule
		Model               & Lr schd & Extra augs & Train flip & Test flip & MS train &  mAP  & AP@$^{0.5}$ & AP@$^{0.75}$ & AR$^{300}$   \\
		\midrule
		Faster-RCNN-r50-fpn & 1x & \xmark & \xmark & \xmark & \cmark & 0.552	&0.912	&0.615	&0.616 \\
		Faster-RCNN-r50-fpn & 1x & \cmark & \xmark & \xmark & \xmark & 0.548	&0.911	&0.608	&0.612\\
		Faster-RCNN-r50-fpn & 2x & \cmark & \cmark & \xmark & \xmark & 0.540	&0.906	&0.596	&0.606 \\
		Faster-RCNN-r50-fpn & 2x & \cmark & \cmark & \cmark & \xmark & 0.510	&0.888	&0.543	&0.584 \\
		\bottomrule
	\end{tabular}
	\label{tab:table4}
\end{table}

In order to obtain even more precise detector Cascade-RCNN \cite{cai2017cascade} was also tested. Image size was reduced for models with ResNeXt-101 in order to keep batch size equal to 2 images. The results are available in Table \ref{tab:table5}. Soft-NMS \cite{bodla2017softnms} was applied as a post-processing, which usually gives better results.

\begin{table}
	\caption{Cascade-RCNN comparison}
	\centering
	\begin{tabular}{lllllllll}
		\toprule
		Model&Img size&Anchor scale& Soft-NMS&mAP& AP@$^{0.5}$ & AP@$^{0.75}$ & AR$^{300}$   \\
		\midrule
		Cascade-RCNN-r50-fpn        &(816, 1088)&[8]& \xmark & 0.525	&0.840 &0.604 &0.582 \\
		Cascade-RCNN-r50-fpn        &(816, 1088)&[4]& \xmark & 0.553	&0.902 &0.626 &0.615 \\
		Cascade-RCNN-r50-fpn        &(816, 1088)&[4]& \cmark & 0.556	&0.900 &0.632 &0.622 \\
		Cascade-RCNN-x101-32x4d-fpn &(768, 1024)&[4]& \xmark & 0.556	&0.903 &0.629 &0.617 \\
		Cascade-RCNN-x101-32x4d-fpn &(768, 1024)&[4]& \cmark & 0.560	&0.902 &0.635 &0.623 \\
		\bottomrule
	\end{tabular}
	\label{tab:table5}
\end{table}

\subsubsection{Tiling strategies}
\label{sec:scale2}
Image tiling as a trick for object detection for large images with small objects on them was previously explored in \cite{ozge2019power}. Following the proposed method 2x2 tiling grid was chosen with 20\% intersection between consecutive tiles. Keeping in mind dense object location bounding box annotations of objects with residual area less than 20\% of the original were removed.
In their work \cite{ozge2019power} proposed to also take a full frame for merging detectors. Although it might seem helpful for "large" objects, I'd like to emphasize that we almost never have those, so this option seems redundant.

Hence, I decided to test the two following approaches: 
\begin{itemize}
	\item \textbf{Tile-combine}. In this scenario each image in train and test parts of the dataset is split into 4 tiles. Appropriate annotations are created. During test time inference is run on each tile and the results are saved into a JSON file. Then the results are manually merged. Duplicated objects on overlapping areas are resolved by NMS. 
	\item \textbf{Train-time tiling}. Here the model on 4 tiles is trained. Test-time inference is run using a 4 times larger resolution. The last one is possible because we do not have to store gradients in the test time. 
\end{itemize}

\emph{Tile-combine} seemed promising when evaluating on tiles (w/o merging), but the metrics dramatically decreased for full frames after combining the results (merged). I also tried to resolve multiple bounding boxes on tiles edges by NMS providing normalized areas instead of probability scores from the detector, which did not improve the score. Obviously when image tiling is applied `anchor\_scale` parameter becomes much less important. Finally, \emph{train-time tiling} and applying detection on a full frame with 1632x2176 resize showed the best results and I used this configuration for final training. See Table \ref{tab:table6}.

\begin{table}[!htp]
	\caption{Tiling strategies}
	\centering
	\begin{tabular}{lllllllll}
		\toprule
		Model                      & Comment & An. scale & Soft-NMS &  mAP  & AP@$^{0.5}$ & AP@$^{0.75}$ & AR$^{300}$   \\
		\midrule
		Faster-RCNN-r50-fpn        & w/o merging & [8] & \xmark & 0.561	&0.912 &0.632 &0.628 \\
		Faster-RCNN-r50-fpn        & w/o merging & [4] & \xmark & 0.566  &0.928 &0.636 &0.636 \\
		Faster-RCNN-r50-fpn        & merged & [4] & \xmark & 0.547	&0.894 &0.615 &0.611 \\
		Faster-RCNN-r50-fpn        & full frame & [4] & \cmark & \textbf{0.577} &0.928 &\textbf{0.659} &\textbf{0.654} \\
		\bottomrule
	\end{tabular}
	\label{tab:table6}
\end{table}

\subsubsection{Final submission approach}

For my final approach I used Faster-RCNN-r50-fpn with `anchor\_scale` = [4] trained on a combination of train, val and test sets tiled as described above. During inference on a provided leaderboard-test set without annotations I used resizing to 1632x2176 and Soft-NMS to resolve duplications. No TTA or ensembling was used.

\section{Conclusions}

In this report it was empirically shown that the simple two-stage detector is a strong baseline for the Product Detection in Densely Packed Scenes task. Also two scale-related approaches are described which improve the detector performance by a large margin. Hence, object scale is one of the aspects that should be considered for further investigation and I also hope to see new approaches being developed for the dataset taking this idea into account.

\bibliographystyle{plain} 
\bibliography{references} 

\end{document}